\documentclass[runningheads]{llncs}
\usepackage[T1]{fontenc}
\usepackage{booktabs}
\usepackage{amssymb}
\usepackage{multirow}
\usepackage{xcolor, colortbl}
\usepackage{tabularx}

\usepackage{todonotes}

\newcommand{\bs}{$\blacksquare$}
\newcommand{\rs}{$\square$}

\usepackage{url}
\usepackage{soul}
\usepackage{xcolor} 
\usepackage{todonotes}
\definecolor{blue}{rgb}{0,0, 0.6}
\definecolor{dkgreen}{rgb}{0,0.6,0}
\definecolor{gray}{rgb}{0.5,0.5,0.5}
\definecolor{mauve}{rgb}{0.58,0,0.82}
\definecolor{mauve}{rgb}{0,0,0}
\definecolor{black}{rgb}{0,0,0}
\definecolor{tri}{rgb}{.25,.88,.82}
\definecolor{lilac}{rgb}{0.85,0.64,0.85}
\definecolor{lightblue}{rgb}{0.53, 0.81, 0.98}
\definecolor{lightskyblue}{rgb}{0.53, 0.81, 0.98}

\begin{document}
\title{The CLEF-2026 FinMMEval Lab: Multilingual and Multimodal Evaluation of Financial AI Systems}

\titlerunning{The CLEF-2026 FinMMEval Lab}
%


\author{%
Zhuohan Xie\inst{1}\orcidID{0009-0008-2650-2857} \and
Rania Elbadry\inst{1}\orcidID{0009-0002-9337-9680} \and 
Fan Zhang\inst{2}\orcidID{0009-0008-7515-3542}	 \and
Georgi Georgiev\inst{3}\orcidID{0009-0006-4418-7622} \and
Xueqing Peng\inst{4}\orcidID{0009-0004-1484-0622} \and
Lingfei Qian\inst{4}\orcidID{0009-0007-8342-1837} \and
Jimin Huang\inst{4}\orcidID{0000-0002-3501-3907} \and
Dimitar Dimitrov\inst{3}\orcidID{0000-0003-1308-180X} \and
Vanshikaa Jani\inst{5}\orcidID{0009-0001-1292-7386} \and
Yuyang Dai\inst{6}\orcidID{0009-0005-4077-0493} \and 
Jiahui Geng\inst{1}\orcidID{0000-0002-4205-8230} \and
Yuxia Wang\inst{6}\orcidID{0000-0002-4474-1826} \and
Ivan Koychev\inst{3}\orcidID{0000-0003-3919-030X} \and
Veselin Stoyanov\inst{1}\orcidID{0009-0003-1250-4820} \and
Preslav Nakov\inst{1}\orcidID{0000-0002-3600-1510} 
}
\authorrunning{Z. Xie et al.}
%
\institute{%
MBZUAI, Abu Dhabi, UAE \and
The University of Tokyo, Tokyo, Japan \and
Sofia University “St. Kliment Ohridski”, Sofia, Bulgaria \and
The Fin AI, USA \and
University of Arizona, Tucson, USA \and
INSAIT, Sofia University “St. Kliment Ohridski”, Sofia, Bulgaria
\\
\email{\{zhuohan.xie, preslav.nakov\}@mbzuai.ac.ae}
}
\vspace{-20mm}
\maketitle              
\begin{abstract}
We present the setup and the tasks of the FinMMEval Lab at CLEF 2026, which introduces the first multilingual and multimodal evaluation framework for financial Large Language Models (LLMs). While recent advances in financial natural language processing have enabled automated analysis of market reports, regulatory documents, and investor communications, existing benchmarks remain largely monolingual, text-only, and limited to narrow subtasks. FinMMEval 2026 addresses this gap by offering three interconnected tasks that span financial understanding, reasoning, and decision-making: Financial Exam Question Answering, Multilingual Financial Question Answering (PolyFiQA), and Financial Decision Making. Together, these tasks provide a comprehensive evaluation suite that measures models’ ability to reason, generalize, and act across diverse languages and modalities. The lab aims to promote the development of robust, transparent, and globally inclusive financial AI systems, with datasets and evaluation resources publicly released to support reproducible research.

\end{abstract}

\section{Introduction}
\label{sec:intro}

The rapid advancement of Large Language Models (LLMs) has propelled significant progress in financial natural language processing (FinNLP), enabling automated analysis of market reports, regulatory documents, and investor communications~\cite{xie-etal-2023-next,multifinben,fire}.
Despite these advances, most existing benchmarks remain monolingual, text-only, and narrowly focused on sub-tasks such as sentiment classification or factual question answering~\cite{finbert,finqa}.
Yet, modern financial communication is increasingly global and multimodal, spanning multilingual news, regulatory filings, and real-time market data. This evolution calls for comprehensive evaluation frameworks that can assess model's ability to reason, generalize, and act across languages and modalities.
To bridge this gap, the FinMMEval Lab at CLEF 2026 extends the evaluation frontier by introducing the first multilingual and multimodal shared tasks for financial LLMs.
Building upon the foundations of earlier initiatives such as Regulations Challenge~\cite{financialregulations}, Fin-DBQA~\cite{findbqa}, and Earnings2Insights~\cite{earnings2insights}, FinMMEval integrates financial reasoning, multilingual understanding, and decision-making into a unified evaluation suite designed to promote robust, transparent, and globally competent financial AI.
It introduces three interconnected tasks  spanning five languages, summarized in Table~\ref{tab:lang}. Each task targets a distinct layer of financial reasoning, from conceptual understanding to real-world decision making, while addressing key challenges in multilingual and multimodal FinNLP:
\begin{itemize}
    \item \emph{Task 1: Financial Exam Question Answering:}
Serves as the foundational component of FinMMEval, focusing on conceptual understanding and domain reasoning (cf.~Section~\ref{sec:task1}).
It evaluates whether models can handle professional exam-style questions across multiple languages, forming the knowledge base upon which the subsequent tasks build.
    \item \emph{Task 2: Multilingual Financial Question Answering (PolyFiQA):}
Extends the conceptual foundation of Task~1 by situating financial reasoning in a multilingual and multimodal context (cf.~Section~\ref{sec:task2}).
It challenges systems to integrate financial reports and news across languages for complex analytical QA, testing their ability to generalize across both linguistic and informational modalities.
    \item \emph{Task 3: Financial Decision Making:}
Synthesizes the knowledge and the reasoning capabilities developed in Tasks~1 and~2 into actionable intelligence (cf.~Section~\ref{sec:task3}).
It assesses reasoning-to-action abilities by requiring models to interpret market dynamics, sentiment, and risk to generate evidence-grounded, risk-aware trading decisions.
\end{itemize}
Together, these tasks establish a multi-layered evaluation framework that bridges understanding, reasoning, and decision-making: three essential pillars for advancing trustworthy, globally capable financial AI systems.


\section{Related Work}
\paragraph{\textbf{Financial AI}}
Financial NLP has advanced from task-specific modeling to large-scale domain adaptation. Early studies such as FinBERT~\cite{finbert} established domain-specific transformers for sentiment and classification, later extending to personal finance~\cite{aipersonalfinance}, credit scoring~\cite{generalistcreditscoring}, and risk-aware evaluation~\cite{yuan-etal-2024-r}. Domain datasets including REFinD, FinARG, FiNER-ORD, and ECTSum~\cite{kaur2023REFinD,ectsum,finer,finben} enabled fine-grained tasks in entity recognition, relation extraction, argument mining, and summarization. Concurrently, English financial NLP benchmarks such as FiQA~\cite{fiqa}, TAT-QA~\cite{tatqa}, FinQA~\cite{finqa}, ConvFinQA~\cite{convfinqa}, FinanceMath~\cite{financemath}, and FinChain~\cite{finchain} have supported progress in sentiment analysis, text–table reasoning, numerical computation, and verifiable reasoning.
Large financial language models have further driven domain progress. BloombergGPT~\cite{wu2023bloomberggpt} demonstrated broad coverage across financial NLP tasks, FinGPT~\cite{liu2023fingpt} emphasized open-source adaptability, and FinMA~\cite{pixiu} achieved competitive performance with compact architectures. Their emergence spurred the creation of benchmarks such as FLANG~\cite{shah-etal-2022-flue}, FinBen~\cite{finben}, and FinMTEB~\cite{finmteb}, which expanded task diversity and standardization. Recent work has also highlighted retrieval and evidence selection as critical bottlenecks in financial QA: approaches such as FINCARDS~\cite{fincards} reformulate document-level QA as constraint-aware evidence reranking, emphasizing explicit alignment over entities, metrics, fiscal periods, and numerical values. More recent evaluations like BizBench~\cite{bizbench} and PIXIU~\cite{pixiu} further revealed persistent weaknesses in quantitative reasoning and multimodal understanding, underscoring the need for models that robustly integrate textual, numerical, and structural signals.
Building on these foundations, multilingual financial NLP has begun addressing global and cross-cultural contexts. Efforts such as CFinBench~\cite{cfinbench}, FLARE-ES~\cite{fit-es}, Plutus~\cite{plutus}, SAHM~\cite{sahm}, and MultiFinBen~\cite{multifinben} extended financial benchmarks beyond English, supporting evaluation across diverse languages and regulatory environments. These initiatives revealed structural disparities in data availability, annotation quality, and task coverage across languages, underscoring the importance of culturally grounded and linguistically inclusive benchmarks for developing globally robust financial LLMs.

\paragraph{\textbf{Related Shared Tasks}}

Given the growing interest in financial NLP, numerous shared tasks have been introduced to address challenges such as misinformation detection~\cite{finlegal25}, financial classification, summarization, stock trading~\cite{finllmssharedtask24}, and causality extraction~\cite{fincausal25}.
While most existing work targets English, emerging efforts have explored low-resource languages such as Arabic~\cite{arafinnlp24}.
Recent initiatives include Earnings2Insights~\cite{earnings2insights}, which evaluates systems on generating persuasive investment guidance from earnings call transcripts; the Regulations Challenge~\cite{financialregulations}, which benchmarks LLMs on understanding and applying financial regulations across nine tasks, and Fin-DBQA~\cite{findbqa}, which assesses multi-turn question answering requiring database querying, multi-hop reasoning, and tabular data manipulation.
Collectively, these shared tasks highlight the increasing breadth and complexity of financial NLP benchmarks, though none has yet focused on multilingual or multimodal financial understanding.

\begin{table}
\centering
\caption{Languages targeted in the three tasks of the CLEF 2026 FinMMEval Lab. White squares indicate languages with test data only.}\label{tab:lang}
\begin{tabularx}{0.95\textwidth}{@{}l*{7}{>{\centering\arraybackslash}X}@{}}
\toprule
 & \textbf{English} & \textbf{Chinese} & \textbf{Arabic} & \textbf{Hindi} & \textbf{Greek} & \textbf{Japanese} & \textbf{Spanish}  \\ \midrule
T1 & \bs & \bs & \bs  & \bs & \rs & & \bs   \\ 
T2 & \bs &  \bs &   &  & \bs & \bs & \bs \\ 
T3 & \bs   \\  
\bottomrule
\end{tabularx}
\end{table}

\section{Task 1: Financial Exam Question Answering }
\label{sec:task1}

\paragraph{Motivation.}
Professional financial qualification exams (e.g., CFA, EFPA) require the integration of theoretical and regulatory knowledge with applied reasoning. 
Existing LLMs often rely on factual recall without demonstrating the analytical rigor expected from human candidates. 
This task evaluates whether models can achieve domain-level understanding and reasoning consistency across multilingual financial contexts.

\paragraph{Task Definition.}
Given a stand-alone multiple-choice question $Q$ with four candidate answers $\{A_1, A_2, A_3, A_4\}$, 
the system must select the correct answer $A^*$. 
The questions cover valuation, accounting, ethics, corporate finance, and regulatory knowledge. 
The focus is on conceptual understanding and precise financial reasoning rather than surface pattern recognition.

\paragraph{Data.}
We combine existing multilingual financial exam datasets with newly collected materials:
\begin{itemize}
    \item \textbf{EFPA}~\cite{multifinben} (Spanish): 230 exam-style multiple-choice questions derived from EFPA certification exam preparation materials, covering investment products, portfolio management, financial regulation, and professional standards. The dataset evaluates applied financial knowledge and conceptual reasoning in Spanish.
    
    \item \textbf{GRFinQA}~\cite{plutus} (Greek): 268 exam-style multiple-choice questions sourced from Greek university-level finance, business, and economics courses. The questions are manually curated and verified by native Greek speakers to ensure linguistic fidelity and domain correctness.
    
    \item \textbf{CFA} (English)~\cite{realfin}: 600 exam-style multiple-choice questions covering nine core domains, including ethics, quantitative methods, financial reporting, corporate finance, and portfolio management. The questions emphasize conceptual understanding and professional-level financial reasoning.
    
    \item \textbf{CPA} (Chinese)~\cite{realfin}: 300 exam-style multiple-choice questions focusing on accounting, auditing, financial management, taxation, economic law, and strategy. The dataset targets domain-specific financial knowledge in Chinese professional certification contexts.
    
    \item \textbf{BBF}~\cite{bhashabench-finance-2025} (Hindi): 500--1000 exam-style multiple-choice questions drawn from over 25 Indian financial and institutional exams, covering problem solving, financial mathematics, governance, and regulatory topics. The dataset reflects the diversity of the Indian financial examination landscape.
    
    \item \textbf{SAHM}~\cite{sahm} (Arabic): 873 exam-style multiple-choice questions from the SAHM benchmark, including accounting and business exam questions derived from authentic Arabic-language examinations. The dataset evaluates financial and business reasoning in Arabic professional and academic contexts.
\end{itemize}
All questions were reviewed by financial professionals to ensure correctness and conceptual balance.

\paragraph{Evaluation.}
The models are required to output the correct answer, and the performance is measured in terms of accuracy, defined as the proportion of correctly identified options in the test set.
Participants may choose to evaluate their systems on any subset of the available languages, including single-language or multilingual settings. Submissions are evaluated within the same task framework, and language coverage is treated as a configurable dimension rather than defining separate subtasks.


\section{Task 2: Multilingual Financial Question Answering}
\label{sec:task2}

\paragraph{Motivation.}
Global finance increasingly demands multilingual reasoning across documents such as English financial filings and foreign-language news. 
However, most existing benchmarks are monolingual. 
This task introduces the first cross-lingual financial reasoning benchmark, testing a model's ability to integrate multilingual textual signals and perform analytical reasoning grounded in authentic financial data.

\paragraph{Task Definition.}
Given a financial report $R$ (English 10-K/10-Q excerpts) and multilingual news articles 
$N = \{N_{en}, N_{zh}, N_{ja}, N_{es}, N_{el}\}$ related to the same company filings, the model receives a question $q$ 
and must generate an answer $a$ supported by multilingual textual evidence. 
Two difficulty tiers are included:
\begin{itemize}
    \item \textbf{PolyFiQA-Easy:} factual or numerical trend questions (e.g., revenue growth, cash flow irregularities);
    \item \textbf{PolyFiQA-Expert:} complex analytical questions requiring multi-document reasoning (e.g., investment strategies, capital allocation).
\end{itemize}

\paragraph{Data.}
We use the PolyFiQA-Easy and PolyFiQA-Expert datasets~\cite{multifinben}, 
which combine U.S. SEC filings with multilingual news (English, Chinese, Japanese, Spanish, Greek). 
Each tier includes 172 QA instances (344 total). 
All questions and answers were manually written and validated by financial experts, with inter-annotator agreement above 89\%. 
The dataset is released under an MIT License.

\paragraph{Evaluation.}
The participating systems are asked to produce concise, evidence-grounded textual answers (up to 100 words). 
We evaluate the performance using ROUGE-1 as the primary evaluation measure. We further calculate BLEURT and factual consistency as secondary measures.

\section{Task 3: Financial Decision Making}
\label{sec:task3}

\paragraph{Motivation.}
Real-world investment decisions require the integration of heterogeneous information sources to form actionable insights, such as textual news, market price dynamics, and current portfolio positions.
Unlike static question--answering tasks, this task centers on \emph{reasoning-to-action}: evaluating how models synthesize complex, time-varying market contexts to generate trading strategies that are both evidence-grounded and risk-aware.

\paragraph{Task Definition.}
Given a market context $C = {P_t, N_t}$, where $P_t$ represents the historical price series, $N_t$ denotes contemporaneous textual information (e.g.,~news, reports), models are required to predict one of three discrete actions: \textit{Buy}, \textit{Hold}, or \textit{Sell}.
Each prediction must be accompanied by a concise textual rationale ($\leq$50 words) that explicitly cites the supporting evidence or reasoning process.

In order to prevent data leakage, the task is set over a pre-specified time window in a live environment. During this period, we will collect each team's daily decisions at a fixed cadence (e.g.,!daily submission by a deadline) based strictly on the information that is available for that day.


\paragraph{Data.}
We use two curated JSON datasets that provide daily market contexts for BTC and TSLA. Our aim is to stress-test LLM trading agents across two materially different market microstructures: (\emph{i})~crypto, which trades 24/7 with fragmented venues, faster regime shifts, and higher tail risk, and (\emph{ii})~equities, which trade in session-based hours with auction opens/closes, corporate-event–driven jumps, and a more regulated market environment.
Each context is indexed by ISO date (YYYY-MM-DD) and aggregates heterogeneous signals for that trading day:

\begin{itemize}
    \item \textbf{Assets and coverage:}
    \begin{itemize}
        \item BTC: Starting from 2025-08-01 and updated daily, the context grows daily (aggregated daily snapshots despite 24/7 trading);
        \item TSLA: Starting from 2025-08-01 and updated daily, the context also grows daily.
    \end{itemize}
\end{itemize}


\begin{itemize}
    \item \textbf{Fields per date $d$:}
    \begin{itemize}
        \item \texttt{prices (float)}: a single representative USD value per day (not OHLC and not intra-day series);
        \item \texttt{news (array[string])}: one or more long-form textual syntheses summarizing that day's market/newsflow;
        \item \texttt{momentum (categorical)}: \{bullish, neutral, bearish\}; a daily market-momentum label manually annotated based on the accompanying news (not a computed technical indicator);
        \item \texttt{future\_price\_diff (float | null)}: this is one-day-ahead price change $P_{d+1} - P_d$; null on the last available date;
        \item \texttt{10k, 10q (array)}: fundamental filings associated with the asset. This component is equity-specific (for TSLA), 10k/10q filings are high-signal fundamental disclosures that update the market’s expectations about earnings power, guidance, and risk. This information is often incorporated rapidly, especially around earnings, these filings can trigger large repricings, with single-day moves on the order of ~10\% not uncommon for some companies.
    \end{itemize}

    \item \textbf{Format:} UTF-8 JSON; keys are calendar dates (no intra-day timestamps).
   
\end{itemize}

This dataset provides the daily market context $C = \{P_t, N_t\}$ for decision-making; models produce actions (Buy/Hold/Sell), while \texttt{future\_price\_diff} serves as the supervisory signal for optimizing your models.

\paragraph{Evaluation.}
We evaluate the model performance for profitability, stability, and risk control using established quantitative metrics:
\begin{itemize}
\item \textbf{Primary:} Cumulative Return (CR);
\item \textbf{Secondary:} Sharpe Ratio (SR), Maximum Drawdown (MD), Daily Volatility (DV), and Annualized Volatility (AV).
\end{itemize}
Together, these indicators capture the model's ability to balance reward and risk, maintain behavioral consistency, and adapt to varying market regimes in a dynamic trading environment.

\section{Conclusion and Future Work}
\label{sec:conclusions}

We presented the design and the setup of the CLEF 2026 FinMMEval Lab, which introduces the first multilingual and multimodal evaluation framework for financial large language models. 
Through three interlinked tasks, including Financial Exam Question Answering, Multilingual Financial QA (PolyFiQA), and Financial Decision Making, the lab advances from conceptual understanding to real-world reasoning and action.
Together, these tasks form a coherent evaluation hierarchy that reflects how financial expertise is built and applied in practice: from knowledge acquisition, to analytical integration across languages and modalities, and finally to evidence-grounded decision-making.

FinMMEval aims to promote transparent, robust, and globally inclusive financial AI systems. By offering multilingual coverage and multimodal contexts, it bridges an important gap in current FinNLP evaluation, moving beyond English-centric and text-only paradigms. The datasets, evaluation code, and baseline models will be publicly released to encourage reproducibility and collaboration across the financial NLP community. 

Future editions will further expand coverage to additional languages, modalities such as charts and regulatory filings, and dynamic real-time evaluation scenarios, paving the way toward trustworthy, adaptive, and globally competent financial LLMs.

\section*{Acknowledgments}
\begin{footnotesize}
We thank all task organizers, and annotators for their valuable contributions to the CLEF 2026 FinMMEval Lab.
We also acknowledge the CLEF community for providing the collaborative platform and infrastructure enabling this multilingual and multimodal evaluation initiative.
The datasets and the evaluation scripts will be publicly released to encourage further research in multilingual and multimodal financial AI.
The work of Dimitar Dimitrov and Ivan Koychev is partially funded by the EU NextGenerationEU project, through the National Recovery and Resilience Plan of the Republic of Bulgaria, project SUMMIT, No BG-RRP-2.004-0008.

\section*{Disclosure of Interests}
The authors have no competing interests to declare that are relevant to the content of this article.

\end{footnotesize}
%
%
%
\bibliographystyle{splncs04}
\bibliography{bib/custom}
\end{document}